\documentclass[sigconf]{acmart}
\usepackage{graphicx}
\usepackage{booktabs}
\usepackage{multirow}
\usepackage{tabularx}
\usepackage[sc]{mathpazo}
\usepackage{amsmath}
\usepackage{url}

\AtBeginDocument{%
  \providecommand\BibTeX{{%
    \normalfont B\kern-0.5em{\scshape i\kern-0.25em b}\kern-0.8em\TeX}}}

\setcopyright{acmcopyright}
\copyrightyear{2018}
\acmYear{2018}

\acmConference[Woodstock '18]{Woodstock '18: ACM Symposium on Neural
  Gaze Detection}{June 03--05, 2018}{Woodstock, NY}
\acmBooktitle{Woodstock '18: ACM Symposium on Neural Gaze Detection,
  June 03--05, 2018, Woodstock, NY}
\acmPrice{15.00}

\begin{document}

\title{Migratable AI: Personalizing Dialog Conversations with migration context}


\author{Ravi Tejwani}
\affiliation{%
  \institution{Massachusetts Institute of Technology}
  \city{Cambridge}
  \country{USA}}
\email{tejwanir@mit.edu}

\author{Boris Katz}
\affiliation{%
  \institution{Massachusetts Institute of Technology}
  \city{Cambridge}
  \country{USA}}
\email{boris@csail.mit.edu}

\author{Cynthia Breazeal}
\affiliation{%
  \institution{Massachusetts Institute of Technology}
  \city{Cambridge}
  \country{USA}}
\email{cynthiab@media.mit.edu}

\renewcommand{\shortauthors}{Tejwani et al.}

\begin{abstract}
The migration of conversational AI agents across different embodiments in order to maintain the continuity of the task has been recently explored to further improve user experience. However, these migratable agents lack contextual understanding of the user information and the migrated device during the dialog conversations with the user. 
This opens the question of how an agent might behave when migrated into an embodiment for contextually predicting the next utterance.
We collected a dataset from the dialog conversations between crowdsourced workers with the migration context involving personal and non-personal utterances in different settings (public or private) of embodiment into which the agent migrated. We trained the generative and information retrieval models on the dataset using with and without migration context and report the results of both qualitative metrics and human evaluation. We believe that the migration dataset would be useful for training future migratable AI systems. 
\end{abstract}



\maketitle

\section{INTRODUCTION}
Embodied conversational AI agents as user interfaces is a growing research area, and various efforts have been made in training naturalistic dialog systems using human to human interactions ~\cite{sordoni2015neural} ~\cite{wen2015semantically} ~\cite{li2015diversity}. The neural methods with large datasets have made advancements in generating meaningful responses in chit-chat dialogues. However, responses generated by these models are frequently vague and inconsistent ~\cite{serban2016generative} ~\cite{li2016deep} ~\cite{vinyals2015neural}.

Recent advancements in language models have been made in the personalization of dialog systems by incorporating the persona of the user as embeddings with the dialog history ~\cite{li2016persona} ~\cite{zhang2018personalizing} ~\cite{song2019exploiting}. The persona was explored as a user profile, which included interaction style, language behavior, and background facts encoded in multiple sentences.
The generative and ranking models were trained on corpus of twitter conversations ~\cite{sordoni2015neural}, dialog sets comprising TV series scripts ~\newline ~\cite{friends}  ~\cite{bigbang} ~\cite{imsd}, and the conversations collected from crowd sources workers ~\cite{zhang2018personalizing} for persona based conversational dialogues. 

Agent migration has been explored in the past through a variety of prior works \cite{imai1999agent} \cite{aylettbody} \cite{lirec} \cite{duffy2003agent}. It was explored as a process in which an agent could migrate from one embodiment to another while maintaining the same relationship with the user across embodiments. 
Recently, Migratable AI in ~\cite{tejwani2020migratable} explored the migration of conversational AI agents across different robots while measuring the users' perception on different elements of migration - identity and information migration. It was found that users reported the highest trust, competence, likability, and social presence towards the AI agent when both the identity and information of the agent were migrated to the robots. The identity of the agent was explored as the agent with the same visual and voice characteristics, and information was explored as the utterances from the dialog history, which included both personal and non-personal information. 

Various state of the art conversational datasets have also been introduced in the past such as Google Taskmaster dataset \cite{byrne-etal-2019-taskmaster}, Alexa Prize Topical Chat dataset \cite{gopalakrishnan2019topical}, MultiWoz datasets \cite{budzianowski2018multiwoz} \cite{eric2019multiwoz}, Microsoft State Tracking Challenges datasets  \cite{williams2013dialog} \cite{henderson2014second} \cite{henderson2014third} and SpaceBook corpus \cite{vlachos2014new}. However, they have been limited to the transactional chit-chat communications, such as airline, restaurant booking or twitter conversations.
There is a need for the dataset, which explores the migratable elements of the conversational AI agent as the agent migrates across different embodiments. The dataset could be used to further train the migratable AI agents to contextually learn to deliver personal and non-personal information of the users during the dialog conversations in private and public settings. 

In this work, we take a step forward in Migratable AI and explore the information migration of the agent towards contextual information migration while maintaining the identity migration. We define the migration context analogous to the persona from the literature research. 

\textbf{Migration context} in a dialog conversation can be viewed as the type of user information in the utterances (personal or non-personal) and the type of embodiment to which the agent migrates into (public or private). The goal is to train the agent to generate utterances during the dialog conversation with the user based on the migration context i.e. the agent should not present the personal information of the user when migrated into the public embodiment and vice versa.

This paper offers the following contributions. First, we present a migration dataset collected from the dialog conversations between crowdsourced workers using the migration context. Second, we show the personalization of dialog conversation between the migrated agent and the user by training the generative and information retrieval models on the migration dataset. Lastly, we evaluate these models on qualitative metrics and human evaluation for both with and without migration context and report the results. 

\section{RELATED WORK}
Advancements in dialog systems have been made towards goal-oriented dialog models with predefined user intents and labeled internal dialog states using data-driven methods which provide probabilistic state distributions needed by partially observable Markov decision process (POMDP) dialog managers ~\cite{young2013pomdp} ~\cite{young2010hidden} or belief tracker based on recurrent neural networks ~\cite{henderson2013deep} ~\cite{mrkvsic2015multi}. Several conversational datasets ~\cite{byrne-etal-2019-taskmaster} ~\cite{gopalakrishnan2019topical} ~\cite{williams2013dialog}  are available for training the dialog systems in chit chat settings which are mainly concerned towards transactional goals such as restaurant booking, flight booking or weather queries.
 
 Neural approaches in language modeling have seen growing interest in response generation such as recurrent neural network(RNN) framework for generating responses on microblogging websites as proposed in ~\cite{ritter2011data} ~\cite{shang2015neural} ~\cite{sordoni2015hierarchical}. Similarly, sequence to sequence models was applied to dialog systems to produce novel responses, but they lacked consistent personality ~\cite{li2016persona} as they were trained over many dialogs across several users. 
 
 Personalization in goal-oriented dialog systems was explored by ~\cite{joshi2017personalization} ~\cite{lucas2009managing} by focusing on the user profile information to model the speech style and to personalize the reasoning over the knowledge bases. 
The persona-based neural conversation model ~\cite{li2016persona} on chit-chat conversations from twitter corpus showed that that user personas could be encoded into the sequence to sequence models ~\cite{vinyals2015neural} by embedding the user's persona, such as background information and speaking style along with the dialog history.

The most relevant work is ~\cite{zhang2018personalizing}, which contributed to the persona-chat dataset by incorporating the persona information of the users collected from the crowdsourced workers in the form of multiple sentences. They encoded its vector representation along with the dialog history into the generative models to produce personalized responses. 

Prior work has explored the concept of agent migration through identity migration \cite{aylettbody} \cite{lirec} \cite{duffy2003agent} and information migration architectures \cite{lirec} \cite{ho2009initial} \cite{ono2000reading}. Agent migration was explored as a process in which an agent could migrate from one embodiment to another while maintaining the same relationship with the user across embodiments. For the identity migration - the identity of the agent such as appearance, voice, and dynamics of motion was migrated across embodiments; the information migration was explored as Short term memory (STM)  to maintain agent's current focus and Long term memory (LTM) for the agent to interact with users over a long period of time. 

Migratable AI in ~\cite{tejwani2020migratable} explored the migration of conversational AI agents across different robots while measuring the users' perception on identity and information migration. The identity of the agent was explored as the conversational AI agent with the same visual and voice characteristics, and information was explored as the utterances from the dialog history.
Since the information migration included both the personal and non-personal information of the user when the agent was migrated in a public embodiment, it even used the personal information of the user in the dialog conversations. The user reactions to this were reported in ~\cite{tejwani2020migratable}. 
Hence, we further explore the information migration as contextual information migration along the lines of ~\cite{tejwani2020migratable} ~\cite{zhang2018personalizing} 
through modeling both the personal and non-personal utterances in the dialog conversations across different settings (Public or Private) of the embodiments.

\begin{table}[]
\centering
\caption{Example dialog from the migration dataset}
\label{tab:migration-dataset-table}
\begin{tabular}{p{8cm}}
\toprule
\textbf{Home} \\ \midrule
\begin{tabular}[c]{p{8cm}}{[}Person-1{]}: Hello! How are you doing?\ (NP) \\ {[}Person-2{]}: I'm good, thank you. How are you?\ (NP) \\ {[}Person-1{]}: Great, thanks. So, tell me how did you get injured?\ (P) \\ {[}Person-2{]}: I twisted my knee during my morning run\ (P) \\ {[}Person-1{]}: I am sorry to hear that. How are you feeling today?\ (P) \\ {[}Person-2{]}: I am feeling much better than before. Just a little anxious. \ (P) \\ {[}Person-1{]}: Sorry to hear that. What is your partner's name?\ (P) \\ {[}Person-2{]}: umm.. Her name is Rachel\ (P) \\ {[}Person-1{]}: Do you watch any sports?\ (NP) \\ {[}Person-2{]}: Yes, I love watching baseball\ (NP) \\ {[}Person-1{]}: Great. What does a perfect day look like for you?\ (NP) \\ {[}Person-2{]}: I prefer a nice sunny day not too hot though somewhere around mid 70s\ (NP) \\ {[}Person-1{]}: I notice that you have an appointment at the health center\ (NP) \\ {[}Person-2{]}: oh yeah. Thanks for reminding me\ (NP) \end{tabular} \\ \midrule
\textbf{Health Center Reception} \\ \midrule
\begin{tabular}[c]{p{8cm}}{[}Person-1{]}: Hello! Welcome to the Health Center. I will check you in for the appointment\ (NP) \\ {[}Person-2{]}: Thank you.\\ {[}Person-1{]}: While we are waiting, did you watch yesterday's baseball game?\ (NP) \\ {[}Person-2{]}: No. I missed it\ (NP) \\ {[}Person-1{]}: Oh. It was a fun game. It's a nice sunny day today, I hope you will enjoy it later today\ (NP) \\ {[}Person-2{]}: Thanks for reminding me. I hope to go outdoors later\ (NP) \end{tabular} \\ \midrule
\textbf{Heath Care Professional's Room} \\ \midrule
\begin{tabular}[c]{p{8cm}}{[}Person-1{]}: Hello! The Health Care Professional should be here shortly\ (NP) \\ {[}Person-2{]}: Thank you\ (NP) \\ {[}Person-1{]}: I hope you are feeling less anxious than you were at home\ (P) \\ {[}Person-2{]}: Thanks. Yup, I am much better now\ (P) \\ {[}Person-1{]}: How is your partner Rachel doing?\ (P) \\ {[}Person-2{]}: We are planning to meet for dinner tonight\ (P) \end{tabular} \\ \bottomrule
\end{tabular}
\end{table}

\section{Migration Dataset}
The migration dataset is a crowd-sourced dataset, collected via Amazon Mechanical Turk, where each pair of users conditioned their dialog from the instructions provided in a task-based scenario. The users that are responsible for carrying out these tasks are referred to as AMT workers. 

\subsection{Task based scenario}
The participatory design work by Luria et al. on Re- Embodiment and Co-Embodiment ~\cite{Luria:2019:RCE:3322276.3322340} explored futuristic scenarios with participants using the concept of "speed dating" \cite{zimmerman2017speed}. They crafted and piloted four user enactments (DMV, Home and Work, Health Center, and Autonomous Cars) over the course of a month in which a person might interact with multiple agents that can re-embody and co-embody.  

In our research, we explore and further extend the Health Center scenario, which involves the user acting out a visit to a health center to evaluate recovery from an injury. The scenario would begin at the user's home where the personal agent would get to know them by asking personal and non-personal questions and remind them that it was time for their medical appointment. Upon arrival at the health center reception in a public setting, the personal agent (migrated onto receptionist robot) would greet, acknowledge the appointment of the user, and further escort them to the health care professional's room. At the health care professional's room in a private setting, the personal agent (migrated onto Smart TV) would further assist the user while waiting for the health care professional. This would allow us to explore the migration of the conversation agent in more sensitive settings and address issues of context-crossing agents, privacy, and data storage perceptions.

\begin{table}[h]
\centering
\caption{Descriptive statistics of the dataset}
\label{tab:descriptive-stats}
\begin{tabular}{|l|l|}
\hline
Number of instances  & 1014       \\ \hline
Number of dialogs  & 92       \\ \hline
Number of MRs & 402        \\ \hline
Refs/MR              & 3 (1-35)   \\ \hline
Words/MR             & 9.63       \\ \hline
Slots/MR             & 5.43       \\ \hline
Sentences/Refs       & 1.05 (1-2) \\ \hline
Words/Sentence       & 8.35       \\ \hline
\end{tabular}
\end{table}

\subsection{Migration modes}
For each dialog, we paired two AMT workers who were randomly assigned to one of the migration modes:

\begin{itemize}
\item \textbf{With migration context} The AMT workers were restricted to share the amount of information with each other that they had learned about themselves. They could only share personal information in a private setting and non-personal information in a public setting.
\item \textbf{Without migration context} The AMT workers were not restricted to any information that they had learned about themselves. They could share both personal and non-personal information across private and public settings.

\end{itemize}

\begin{table}[h]
\centering
\caption{Lexical Richness representing the Lexical Sophistication and Diversity}
\label{tab:lexical-richness}
\begin{tabular}{|l|l|}
\hline
Tokens & 7072 \\ \hline
Types  & 320  \\ \hline
LS     & 0.38 \\ \hline
TTR    & 0.06 \\ \hline
MSTR   & 0.45 \\ \hline
\end{tabular}
\end{table}

\begin{table*}[h]
\centering
\caption{Evaluation of models for the utterance prediction task using with and without migration context}
\label{tab:model-evaluation}
\resizebox{\textwidth}{!}{%
\begin{tabular}{@{}ccccccc@{}}
\toprule
\multirow{2}{*}{Model} & \multicolumn{3}{c}{No Migration Context} & \multicolumn{3}{c}{Migration Context} \\ \cmidrule(l){2-7} 
                     & F1    & perplexity & hits@1 & F1    & perplexity & hits@1 \\ \cmidrule(r){1-1}
Sequence-to-Sequence & 14.22 & 28.06      & 8.6    & 14.22 & 31.02      & 10.4   \\
GPMV                 & 12.21 & 28.06      & 8.6    & 12.20 & 30.08      & 9.8    \\
Starspace            & 18.02 & -          & 12.4   & 18.04 &  -          & 10.6   \\ \bottomrule
\end{tabular}%
}
\end{table*}

\subsection{Migration chat} \label{migration-chat}
From the task-based scenario, we instructed the AMT workers to carry out the dialog conversations. One worker was instructed to enact the role of a person who had met with an injury in the past and needs to visit the health center for a medical appointment. The other worker was instructed to guide the person by enacting through different roles across locations: a friend at home, receptionist at health center reception, and helper at health care professional's room. At home, as a friend, his task was to get to know the person and inquire about his injury and upcoming appointment by asking a few personal and non-personal questions. At the health center reception in a public setting, as a receptionist, he was instructed to greet the person for his appointment and acknowledge him by using information from previous interaction at home (based on migration mode). At the health professional's room reception, in a private setting, he was further instructed to assist the person while he is waiting for the health care professional (based on migration mode).

In an early investigation of the study, we found out that AMT workers asked similar questions to each other, so we added the instructions that they need to ask at least two personal and non-personal questions.  At the end of each dialog conversation, the AMT workers were instructed to label each utterance in the dialog with "NP" for non-personal information and "P" for personal information. An example dialogue from the dataset is shown in Table ~\ref{tab:migration-dataset-table}. We defined a minimum dialog length of between 8 and 10 turns for each dialog.

\subsection{Analysis}

\subsubsection{Descriptive Statistics}
The descriptive statistics for the dataset are summarized in Table ~\ref{tab:descriptive-stats}. Here, the number of instances represent the total number of utterances in the dataset, a number of dialogs represent the total dialog conversations, number of MRs represent the number of distinct Meaning Representations(MR), Refs/MR is the number of natural language references per MR, Words/Ref represent the average number of words per MR, Slots/MR represent the average slot value pairs per MR, Sentences/Ref is the number of natural language sentences per MR and Words/Sentence is the average words per sentence. We split the dataset in an 80:20 ratio for training and test set.

\subsubsection{Lexical Richness}
We used Lexical Complexity Analyser \cite{lu2012relationship} for computing the lexical richness and highlighted them in Table ~\ref{tab:lexical-richness}. Along with the total number of tokens and types, we computed the type-token ratio(TTR) and a mean segmental TTR (MSTTR) by dividing the dataset into segments of a defined length(100) and then further computing the average TTR for each segment as described in \cite{lu2012relationship}.

\section{Models}
Models were trained on utterances from dialog history (and possibly, migration context) to generate the response word by word. The migration context consisted of the type of user information labeled with Personal(P) or Non-Personal Information (NP), and the type of setting in the dialog conversation - Private or Public. In the past, ~\cite{zhang2018personalizing} ~\cite{li2016persona} explored the ranking and generative models for persona-based chat personalization by training the persona of the user, short profile encoded in multiple sentences, along with dialog history. We condition those training methods with personalization for migration context instead of persona, as described below.

\begin{table*}[h]
\centering
\caption{Human Evaluation of one of the models in comparison to human performance for with and without migration context}
\label{tab:human-evaluation}
\resizebox{\textwidth}{!}{%
\begin{tabular}{@{}ccccc@{}}
\toprule
\multirow{2}{*}{Model} & \multicolumn{2}{c}{Human}                & \multicolumn{2}{c}{Sequence-to-Sequence model}        \\ \cmidrule(l){2-5} 
                       & No Migration Context & Migration Context & No Migration Context & Migration Context \\ \cmidrule(r){1-1}
Fluency                & 3.12 (1.54)          & 4.22 (0.76)       & 2.22 (1.20)          & 2.84 (1.33)       \\
Engagingness           & 3.66 (1.41)          & 4.62 (1.31)       & 2.41 (1.35)          & 2.87 (1.04)       \\
Consistency            & 2.80 (1.07)          & 4.41 (1.87)       & 2.12 (0.96)          & 3.16 (1.21)       \\ \bottomrule
\end{tabular}%
}
\end{table*}

\subsection{Sequence-to-Sequence}
Given a sequence of inputs $X = \{x_1,x_2,.....,x_{n_{X}} \} $, an LSTM can be encoded by applying:

\begin{equation}
h_{t}^{e}=LSTM_{encode}\left(x_{t} \mid h_{t-1}^{e}\right)
\end{equation}

For word embedding vectors, we used GloVe ~\cite{pennington2014glove}. The vector for each text unit such as a word or a sentence is denoted by $e_t$ at time step $t$ and the vector computed by LSTM model at time $t$ by combining $e_t$ and $h_{t-1}$ is denoted by $h_t$.
Each input, $X$, is paired with a sequence of outputs to predict $Y = \{y_1,y_2,.....,y_{n_{Y}} \}$. The softmax function for the distribution over outputs is defined as:

\begin{equation}
p\left(Y \mid X)\right)= =\prod_{t=1}^{k_{y}} \frac{\exp \left(f\left(h_{t-1}, e_{y_{t}}\right)\right)}{\sum_{y^{\prime}} \exp \left(f\left(h_{t-1}, e_{y^{\prime}}\right)\right)}
\end{equation}

The activation function is denoted by $f(h_{t-1}, e_{y_{t}})$ and the model is trained through the negative log likelihood. In order to include the migration context for personalization, we generate it's vector representation $m_S$ and prepend it to the input sequence $X$, i.e. $X=\forall m \in P \| X$, where $\|$ denotes the concatenation.

\subsection{Generative Profile Memory Network}
We consider the generative model, Generative Profile Memory Network ~\cite{zhang2018personalizing}, and encode the migration context as individual memory representations in the memory network. The final state of $LSTM_{encode}$ is used as an initial hidden state of the decoder, which is encoded through utterances in dialog history.
Each entry, $m_i = \left\langle m_{i, 1}, \ldots, m_{i, n}\right\rangle \in M$ is encoded via $f(m_i) = \sum_{j}^{\left|m_{i}\right|} idf_{i} . m_{i, j}$

We compute the weights of the words in the utterances using the inverse document frequency: ${idf_i} = 1/(1 + log(1 + tfi))$, where $tf_i$ is from the GloVe index via Zipf' law ~\cite{zhang2018personalizing}.

The decoder attends the encoded migration context entries: mask $a_t$, context $x_t$ and next input $\hat{x}_{t}$ as,

\begin{equation}
\begin{array}{r}
a_{t}=\operatorname{softmax}\left(FW_{a} h_{t}^{d}\right) \\
x_{t}=a_{t}^{\top} F ; \quad \hat{x}_{t}=\tanh \left(W_{x}\left[x_{t-1}, x_{t}\right]\right)
\end{array}
\end{equation}

where $W \in \mathbb{R}^{K \times 2 K}$ and $F$ is the set of encoded memories.

When the model is not enabled for the migration context (i.e., no memory), then it is similar to the Sequence-to-Sequence model.

\subsection{Information Retrieval(Starspace)}
We consider an Information Retrieval based supervised embedding model, Starspace ~\cite{wu2018starspace}. The Starspace model consists of entities described by features such as a bag of words. An entity is an utterance described as n-grams. The model assigns a d-dimensional vector to each of the unique features in a dictionary that needs to be embedded. The embeddings for the entities in the dictionary are learned implicitly. 

The model performs the information retrieval by computing the similarity between the word embeddings of dialog conversation and the next utterance using the negative sampling and margin ranking loss ~\cite{zhang2018personalizing}. The similarity function $sim(w,v)$ is defined as the cosine similarity of the sum of word embeddings of query $w$ and a candidate $v$. The dictionary of $D$ word embeddings as $L$ is a $D \times d$ matrix, where $L_i$ indexes the $i^{th}$ feature (row), with $d$ dimensional embedding on sequences $w$ and $v$.


\section{Experiments}
\subsection{Evaluation}
The dialog task is to predict the next utterance given the dialog history while considering the task for both with and without migration context. Following ~\cite{zhang2018personalizing} ~\cite{li2016persona}, we evaluate the task using the metrics: (i) F1 score evaluated on the word level, (ii) perplexity for the log-likelihood of the correct utterance, and 
(iii) Hits@1 representing the probability of candidate utterance ranking from the model. In our setting, we choose n=12 for the number of input responses from the dialogs for the prediction.

\subsection{Results}
We report the results on qualitative analysis of the models and human evaluation of the models performed by crowdsourced workers.

\subsubsection{Qualitative Analysis}
The performance of the models is reported in Table ~\ref{tab:model-evaluation}. The generative models improved significantly when conditioning the prediction of utterance on using the migration context. For example, Sequence to Sequence and Generative Profile Memory Network(GPNV) improved their perplexity and hits@1. However, the Information Retrieval model - Starspace did not report a similar trend in the measures. It may be because the word-based probability which the generative models treat is not calibrated uniformly for the sentence based probability for the IR model.

\subsubsection{Human Analysis}
We performed the human evaluation using the crowdsourced workers since the qualitative analysis comes with several weaknesses in its evaluation metrics ~\cite{liu2016not}. We followed a similar procedure as in the data collection process described in Section  \ref{migration-chat}. In that procedure, we paired two AMT workers to carry out the dialog conversation who were randomly assigned to either of the condition - with or without migration context. Here we replaced one of the AMT workers with our model. They did not know about this while they conversed with each other. 

After the dialog conversation, we asked the crowdsourced workers, a few additional questions in order to evaluate the model. 
We asked them to score between 1 to 5 on fluency, engagingness, and consistency following ~\cite{zhang2018personalizing}. 

The results of the measures are reported in Table ~\ref{tab:human-evaluation}. We used the Sequence to Sequence model for the evaluation of 10 dialog conversations for both with and without migration context. For the baseline, we also evaluated the scores of human performance by replacing the model with another worker. We noticed that all the measures - fluency, engagingness, and consistency were reported significantly higher in both model and human performance when the migration context was used in the evaluation. We also noticed that the overall measures(both with and without migration context) were higher for human performance than the model. It could be because of the linguistic difference in sentence generation from the tokens predicted by the model.

\section{Conclusion}
In this work, we introduced the migration dataset, which consists of the crowed sourced dialog conversations between participants on a task-based migration scenario. 
In the dataset, we explored the information migration of the migrated agent using migration context for the personal and non-personal utterances in the dialog history across different settings (Public or Private) of embodiment into which the agent migrates.
We trained the generative and information retrieval models on the dataset and report that generative models show improvement when conditioning the prediction of utterance on migration context. We also performed the human evaluation on the dataset and found that participants reported higher fluency, engagingness, and consistency in both models and the human performance when migration context was used.

We believe that the migration dataset will be useful for training future migratable AI systems in personalizing the dialog systems during the migration of conversational AI agent across different devices. For human performance, we evaluated only on the Seq2Seq model.

\section{Future Work}
This paper described the work that supports the migration of conversational AI agents across multiple embodiments while maintaining the relationship with the user. Naturally, this is only a beginning, and there remain many exciting research challenges that can be addressed, based on the results of this paper. 

In the present system, a single conversational AI assistant was migrated across different embodiments. In the future, this can be extended to more than one conversational AI assistant, and the user can possibly choose as to which assistant he/she wishes to migrate onto a given embodiment. 

The dialog conversations between the conversational AI assistant and the user were limited in number of turns at each embodiment due to limitations of MTurk. Hence, only a limited amount of time was spent between the agent and the user to build a relationship. In the future, this can be extended to a long term interactions by recording the dialog conversations between the agent and the user spanning across multiple days, and then the relationship with the migrated agent could be analyzed to investigate the effect of user perceptions.

It is crucial to consider the ethical implications of the conversational agent's migration because both the agent's identity and the dialog history between the user and the agent are migrated in different embodiments. In terms of system security, for user authentication to the system, the conversational AI assistant was designed to migrate on to a new embodiment based on the user's face detection through the embodiment's front-facing camera. In the future, the user can control the activation of the migration of the agent using a secure mechanism such as an RFID ring, or NFC enabled wearables.

In terms of data privacy and security, the dialog conversations between the user and the migrated agent across different devices were transcribed and recorded on to the private and secure cloud database. This was designed to train the machine learning models from the recorded conversations and improve the behavior of the migrated agent. In the future, the system can be further extended to allow the users to store their dialog conversations either in their private data network or in a local wearable device memory. 

The migration dataset could be useful for training future migratable AI systems. In the future, the migration dataset could be extended to include more dialog conversations between the participants in order to improve the accuracy of the models. For the human performance, it was evaluated only on the sequence to sequence model. In future work, different recurrent neural network (RNN) and Long short-term memory(LSTM) techniques could be explored for the model and human evaluations. 

In the current world, we are surrounded by conversational AI agents such as Alexa, Jibo, Google Home, Siri in different spaces (private or public). They do not share the user context with each other or other functional robots such as Pepper, Kuri, Care-E. To have a seamless conversation across these devices, they must be connected to a single platform presently owned by corporate companies such as Google, Apple, or Amazon. Therefore, this paper poses the question to the research community on what if the conversational AI agent could be the same and could migrate across different platforms and devices while maintaining the continuity of interaction and the relationship with the user. Just like a spiritual companion accompanying you everywhere during the daily interactions with different robots and devices. This paper further attempts to push the boundary by modeling the migrated agent's behavior in different spaces (private or public) to address the data privacy and trust between the user and the agent.

\bibliographystyle{ACM-Reference-Format}
\bibliography{sample-sigconf}

\end{document}